\documentclass[letterpaper]{article} 
\usepackage{aaai24}  
\usepackage{times}  
\usepackage{helvet}  
\usepackage{courier}  
\usepackage[hyphens]{url}  
\usepackage{graphicx} 
\usepackage{xcolor}
\usepackage{subfigure}
\usepackage{amsfonts}
\usepackage{natbib}  
\usepackage{caption} 
\usepackage{algorithm}
\usepackage{algorithmic}

\usepackage{newfloat}
\usepackage{listings}
\DeclareCaptionStyle{ruled}{labelfont=normalfont,labelsep=colon,strut=off} 
\floatstyle{ruled}
\newfloat{listing}{tb}{lst}{}
\floatname{listing}{Listing}
\title{APLA: Additional Perturbation for Latent Noise with Adversarial Training Enables Consistency}
\author {
    Yupu Yao\textsuperscript{\rm 1},
    Shangqi Deng\textsuperscript{\rm 1},
    Zihan Cao\textsuperscript{\rm 1},
    Harry Zhang\textsuperscript{\rm 2},
    Liang-Jian Deng\textsuperscript{\rm 1}
}
\affiliations {
    \textsuperscript{\rm 1}University of Electronic Science and Technology of China, Chengdu, China\\
    \textsuperscript{\rm 2}Massachusetts Institute of Technology, Cambridge, USA\\
    yypseek123@gmail.com, dengsq5856@126.com, dengsq5856@126.com,harryz@mit.edu, liangjian.deng@uestc.edu.cn
}

\usepackage{bibentry}

\begin{document}

\maketitle
\begin{abstract}
Diffusion models have exhibited promising progress in video generation. However, they often struggle to retain consistent details within local regions across frames. One underlying cause is that traditional diffusion models approximate Gaussian noise distribution by utilizing predictive noise, without fully accounting for the impact of inherent information within the input itself. Additionally, these models emphasize the distinction between predictions and references, neglecting information intrinsic to the videos. To address this limitation, inspired by the self-attention mechanism, we propose a novel text-to-video (T2V) generation network structure based on diffusion models, dubbed Additional Perturbation for Latent noise with Adversarial training (APLA). 
Our approach only necessitates a single video as input and builds upon pre-trained stable diffusion networks. Notably, we introduce an additional compact network, known as the Video Generation Transformer (VGT). This auxiliary component is designed to extract perturbations from the inherent information contained within the input, thereby refining inconsistent pixels during temporal predictions. We leverage a hybrid architecture of transformers and convolutions to compensate for temporal intricacies, enhancing consistency between different frames within the video. Experiments demonstrate a noticeable improvement in the consistency of the generated videos both qualitatively and quantitatively.
\end{abstract}

\section{Introduction}
Generating video is a challenging task in computer vision, whose aim is to generate high-fidelity, diverse videos from various inputs like text, images, audio, or sketches. Recent works in deep learning have spurred considerable advancements in this domain, most notably through the advent of diffusion models \cite{ho2020denoising}. These models craft videos by iteratively introducing noise to an initial input, then undoing this noise through a series of denoising stages. Their strength is particularly notable in generating high-definition, extended-duration videos with intricate semantics and dynamics.

\begin{figure}
  \centering
  \subfigure[Tune-A-Video \cite{wu2022tune}]{\includegraphics[width=\linewidth]{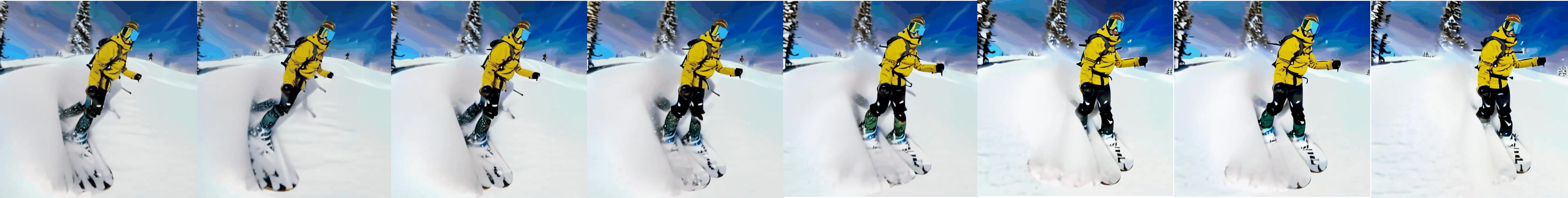} \label{reco}}
  \subfigure[APLA (Ours)]{\includegraphics[width=\linewidth]{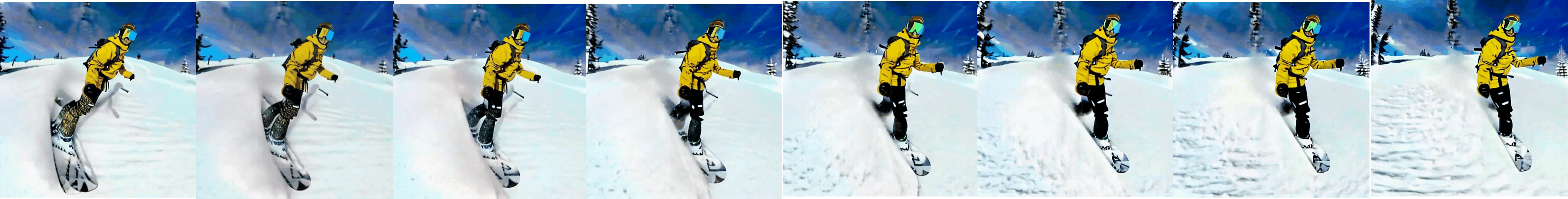} \label{reca}}
  \caption{\textbf{The comparison (by the same prompt: ``A man is skiing'')  between Tune-A-Video and the proposed APLA.} (a) The result of Tune-A-Video is that the snowboard splits into multiple parts on these frames. (b) The obtained outcome is by our APLA method which keeps the single snowboard in all frames.}
  \label{COMP}
\end{figure}

With the advancing capabilities of diffusion models, cross-modality generation tasks, including text-to-image (T2I) and text-to-video (T2V), have made substantial progress. Yet, the substantial data size of videos presents challenges in training video generation models from scratch. A recent approach, ``Tune-A-Video'' \cite{wu2022tune}, aimed to utilize pre-trained T2I models for video synthesis. However, its outcomes exhibited inconsistencies across video frames. Ensuring frame consistency in video generation, especially within fine-tuned models, remains a hurdle. While some efforts have achieved acceptable frame consistency using diffusion models \cite{hoppe2022diffusion}, intricate video details, especially in complex scenarios, are often absent. For instance, even with identical prompts and inputs, the ``Tune-A-Vide'' approach still manifests inconsistencies in generated videos. Moreover, extending fine-tuning epochs could potentially compromise the semantic coherence between successive frames.

\begin{figure}
  \centering
  \subfigure[An SUV is moving on the road, cartoon style]{\includegraphics[width=0.45\textwidth]{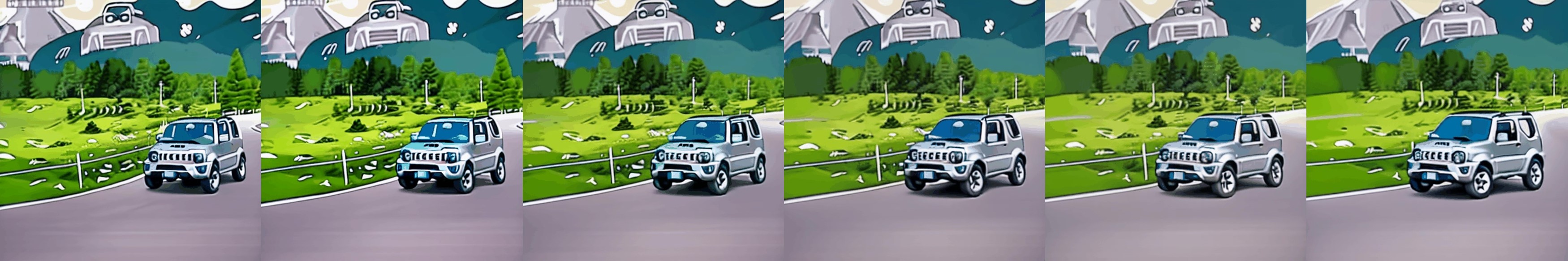} \label{1}}
  \subfigure[An SUV is moving on the beach]{\includegraphics[width=0.45\textwidth]{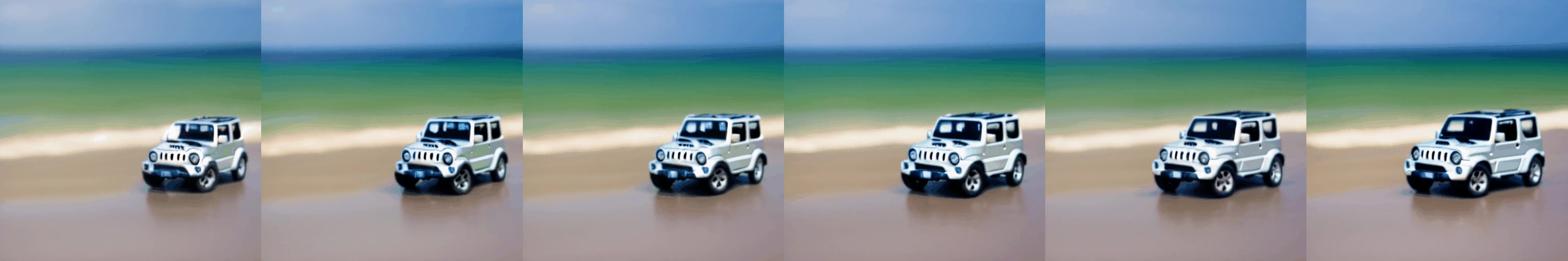} \label{2}}
  \subfigure[A puppy is eating an orange]{\includegraphics[width=0.45\textwidth]{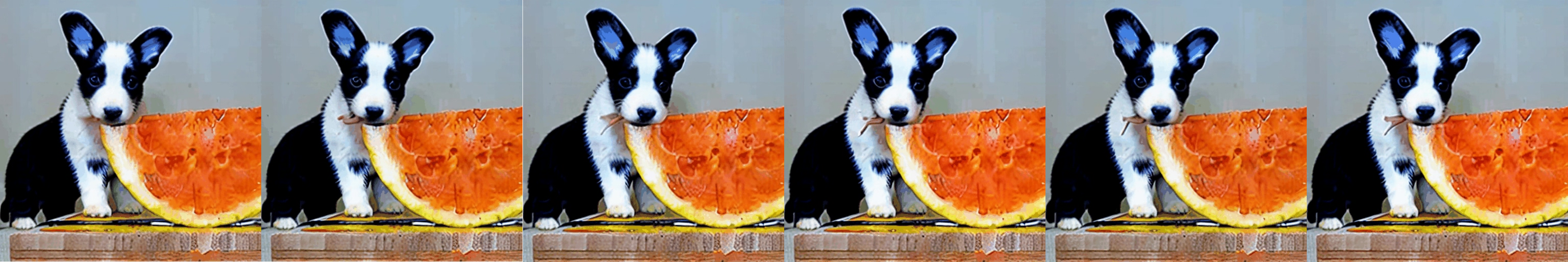} \label{3}}
  \subfigure[A rabbit is eating an orange]{\includegraphics[width=0.45\textwidth]{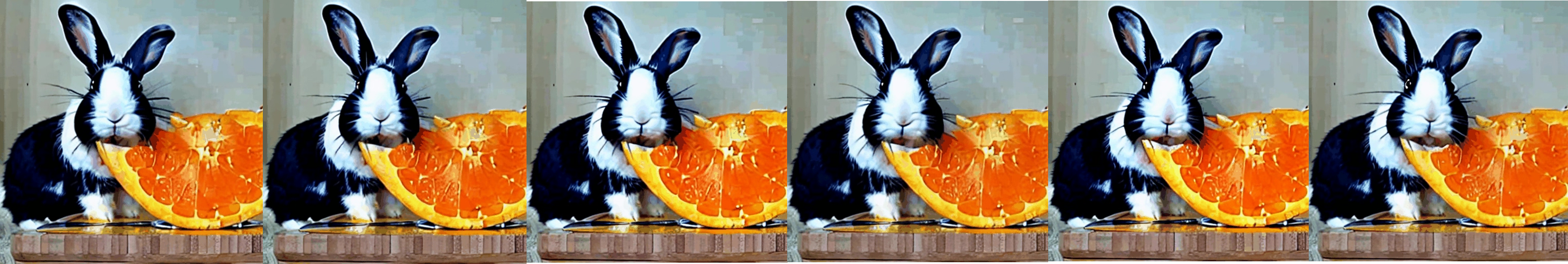} \label{4}}
  \caption{Visual demonstrations of APLA using different prompts.}
  \label{result}
\end{figure}

In light of these challenges, we propose a novel architecture, \emph{i.e.}, APLA. Using self-attention \cite{vaswani2023attention}, APLA is geared to capture inherent video characteristics and establish connections between frames by adaptively generating parameters. To facilitate efficient information extraction from inputs, we devise a much smaller model compared to diffusion models, as shown in the success of style transfer and multi-task learning  \cite{zhang2023adding}. The result can be observed visually in Fig.~\ref{result}. Consequently, we engineer a decoder-only structure for our Video Generation Transformer (VGT). With masking and self-attention mechanisms, VGT exhibits an improvement in predicting unknown frames, demonstrating its ability to distill intrinsic input information. We also introduce a novel loss function, hyper-loss, to encourage the model to focus on nuanced input details. Lastly, to further enhance the generated quality of video and improve the consistency between different frames, we introduce adversarial training to improve the quality of output while strengthening the robustness. The contributions of this work can be summarized as follows:
\begin{enumerate}
    \item A novel architecture, \emph{i.e.}, VGT, builds on top of pre-trained diffusion models, which enhances the consistency between video frames by learning the correlation information between input frames.

    \item A fusion of the diffusion model and adversarial is employed for video generation, where adversarial training is directly applied to the discrepancy in noise distributions, rather than judging the similarity between input and output images.

    \item Quantitative and qualitative experiments that demonstrate the effectiveness of the proposed approach, which achieves SOTA performance in frame consistency of generated videos.
\end{enumerate}

\section{Related Work}
\subsection{Cross-Modal Video Generation}
Generating videos using multiple input modalities presents a formidable challenge within the realm of deep learning. A particularly prominent endeavor in video generation is the synthesis of videos through text-to-video (T2V) techniques, which entails creating videos grounded in natural language descriptions. This innovative synthesis process can be conceptualized as an evolution beyond the established domain of text-to-image (T2I) synthesis. In a broader context, models for text-to-image (T2I) synthesis can be systematically categorized into two distinctive classes: transformer-based models and diffusion-based models. The former category, as exemplified by references such as \cite{ramesh2021zeroshot, sim2019personalization, zhang2016health}, leverages the power of extensively trained large-scale language models such as GPT-3 or T5. These models adeptly transform textual input into latent vectors, which are subsequently employed for downstream image generation. Conversely, the latter category represented by models like \cite{nichol2022glide, zhang2020dex, zhang2021robots}, also incorporates a text encoder in a similar vein, yet diverges in their approach by integrating the text-encoded information into the diffusion process. The prowess of diffusion-based models in crafting intricate images stands evident. This capability has been the foundation for their evolution from text-to-image (T2I) synthesis to the more dynamic realm of text-to-video (T2V) synthesis. However, these initial methods had shortcomings like detail deficiency, temporal inconsistencies, and limited control. Recent approaches have emerged to address these issues and enhance text-to-video (T2V) synthesis.

\subsection{Diffusion Model}
The inception of the Denoising Diffusion Probabilistic Model (DDPM), documented in \cite{ho2020denoising}, took inspiration from thermodynamic frameworks. This model, functioning as a Markov chain, exhibits an auto-encoder structure. Nevertheless, its inference process faced sluggishness owing to the intricate denoising stages encompassed. To address this issue, Denoising Diffusion Implicit Models (DDIM)\cite{song2020denoising} introduced an innovative strategy: by introducing variable variance for predicted noise, they achieved swift inference for the diffusion model within a concise span of steps. Building upon the foundations laid by DDIM, subsequent endeavors \cite{watson2022learning, zhang2023flowbot++} aimed to propel DDPM inference to even greater speeds.

\begin{figure*}[!t]
  \centering
  \includegraphics[width=0.8\textwidth]{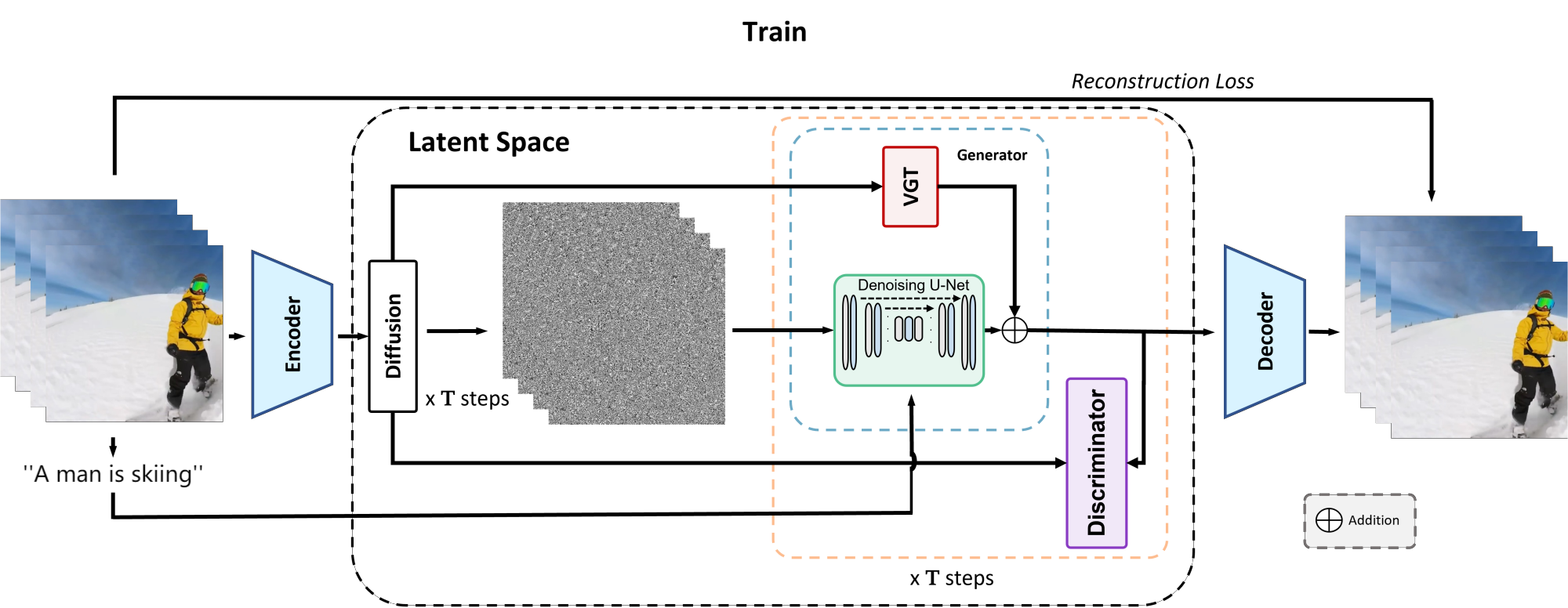}
  \caption{\textbf{The process of training the networks.} VGT extracts intrinsic information from latent variables, considering various time steps for noise incorporation, and especially including the clean latent variable devoid of noise, namely the original latent variable $z$. As VGT is not trained ever, the output of VGT is tiny thus the change of the output is small, which is helpful to improve the consistency of different frames without changing the content a lot. The discriminator receives the predicted noise and the noise residuals for corresponding time steps in the diffusion stage.}
  \label{train}
\end{figure*}

Broadly speaking, the Diffusion model's key strength lies in its remarkable capacity for handling tasks involving cross-modality and multi-modality interactions. This strength is vividly demonstrated by the Latent Diffusion Model (LDM) \cite{rombach2022high, pan2023tax, eisner2022flowbot3d}, which not only exhibited the potential of diffusion for high-resolution image generation through the utilization of latent spaces but also showcased the model's ability to excel in cross-modality scenarios. The application of the Diffusion model to video generation has prompted various approaches \cite{ho2022imagen, lim2021planar, lim2022real2sim2real, avigal20206, avigal2021avplug}. Among these, Tune-A-Video \cite{wu2022tune, devgon2020orienting, shen2024diffclip, jin2024multi} stands out for introducing a novel perspective on video generation via diffusion modeling. This innovative methodology views video generation as a process of refining a pre-trained stable diffusion. In doing so, Tune-A-Video reshapes our understanding of how diffusion models can be harnessed to address the challenges of video generation with a fresh and effective approach.


\section{Methodology}
In this section, we begin by presenting the overall structure of APLA. Subsequently, we delve into the details of VGT, designed to extract intrinsic information. Notably, we introduce two versions of VGT, each showcasing distinct advantages in the experiment. Following this, we discuss hyper-loss and introduce our adversarial training strategy.

\subsection{APLA}
To enhance inter-frame consistency, alongside optimizing high-quality outputs, it's imperative to account for the interconnections among distinct frames. While prior research assumed that inherent information could be naturally grasped by the model without supplementary steps, the challenge becomes more pronounced in complex tasks like video generation, where depicting high-level temporal features during inference proves more intricate compared to image generation. In light of this, we introduce a novel architecture (depicted in Fig.~\ref{train}) that builds upon the diffusion model and integrates an additional module. This module is specifically designed to capture intrinsic information and foster inter-frame connections within the temporal domain. This approach sets us apart from previous methods and addresses the nuances presented by video generation. A visual comparison between APLA and Tune-A-Video is illustrated in Fig.~\ref{COMP}. Specifically, the added module incorporates a self-attention mechanism aimed at extracting information directly from inputs, all without introducing any additional loss. Our hypothesis is rooted in the potential of self-attention mechanisms to gather relevant details from the input itself. This implies the ability to dynamically generate parameters based on input, empowering the model with strong inductive capabilities.

\begin{algorithm}[!t]
    \caption{Additional Perturbation for Latent Noise with Adversarial Training (APLA)}
    \label{APLA}
    \begin{algorithmic}
    \renewcommand{\algorithmicrequire}{\textbf{Input:}}
    \renewcommand{\algorithmicensure}{\textbf{Output:}}
        \REQUIRE{A text query $q$, a reference video $v$, a pre-trained T2I diffusion model with encoder $\varepsilon$ and decoder $\mathcal{D}$, a number of iterations $T$, VGT $\varphi$, a discriminator $D$}
        \ENSURE {A generated video $\hat{x}$ that matches the text query}
        \STATE {Initialize $\hat{x}$ as an empty list}
        \STATE {Extract the first frame $x_{0}$ from $v$}
        \STATE {Encode $x_{0}$ and $q$ into a latent code $z_{0}$ using DDIM inversion}
        \FOR{$t=1$ to $T$}
        \STATE {Compute $z_{t+1}$ through Eq.~\ref{diff}}
        \STATE {Fine-tune the model parameters $\theta$ of $\varepsilon$ using Eq.~\ref{final}}
        \ENDFOR
        \FOR{$t=T,T-1,...,1$}
        \STATE {Sampling $\hat{z}_{t-1}$ using DDIM inversion}
        \IF{$t==1$}
        \STATE {Obtain $\hat{x}=\mathcal{D}(\hat{z}_{t})$}
        \ENDIF
        \ENDFOR
        \RETURN $\hat{x}$
    \end{algorithmic}
\end{algorithm}
\begin{figure*}[!t]
  \centering
  \includegraphics[width=\textwidth]{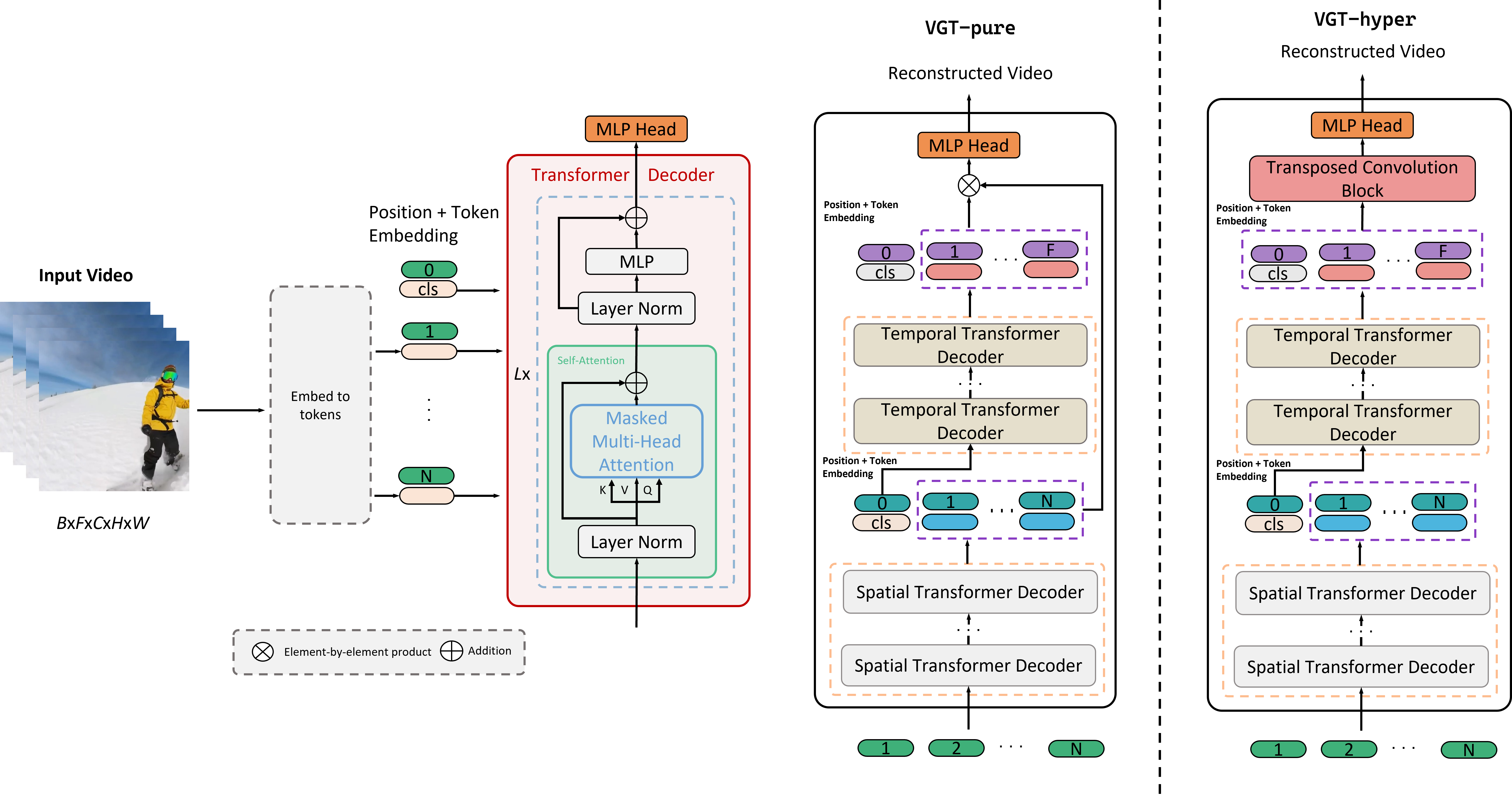}
  \caption{\textbf{An illustration of VGT-Pure and VGT-Hyper.} The left side shows the transformer decoder structure, which adapted mask operation on the self-attention mechanism especially. The right side shows the two versions of VGT. The Temporal Transformer Decoder only receives the class (\emph{i.e.}, \textit{cls}) token of output sequences of the Spatial Transformer Decoder. The rest of the tokens of the output of the Spatial Transformer Decoder are used to multiply with the tokens of the Temporal Transformer Decoder output dislodging \textit{cls} token in VGT-Pure, while the whole output of the Temporal Transformer Decoder is transmitted to Transposed Convolution Block directly in VGT-Hyper.}
  \label{VGT}
\end{figure*}

\begin{figure*}[!t]
  \centering
  \subfigure[Comparison of VGT versions]{\includegraphics[width=0.45\textwidth]{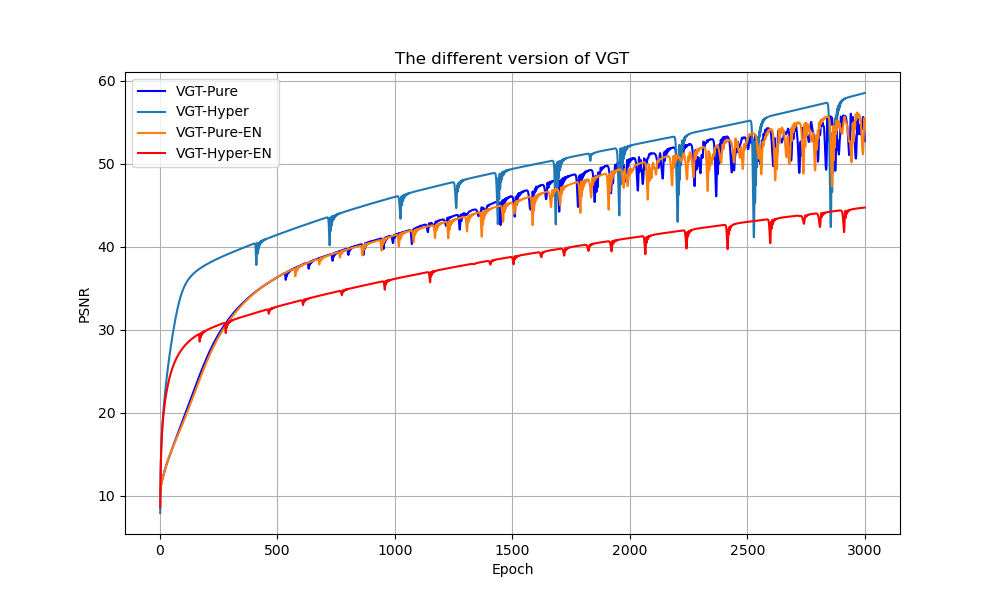} \label{VGT_COM}}
  \hfill
  \subfigure[Ratio of VGT output and U-Net]{\includegraphics[width=0.45\textwidth]{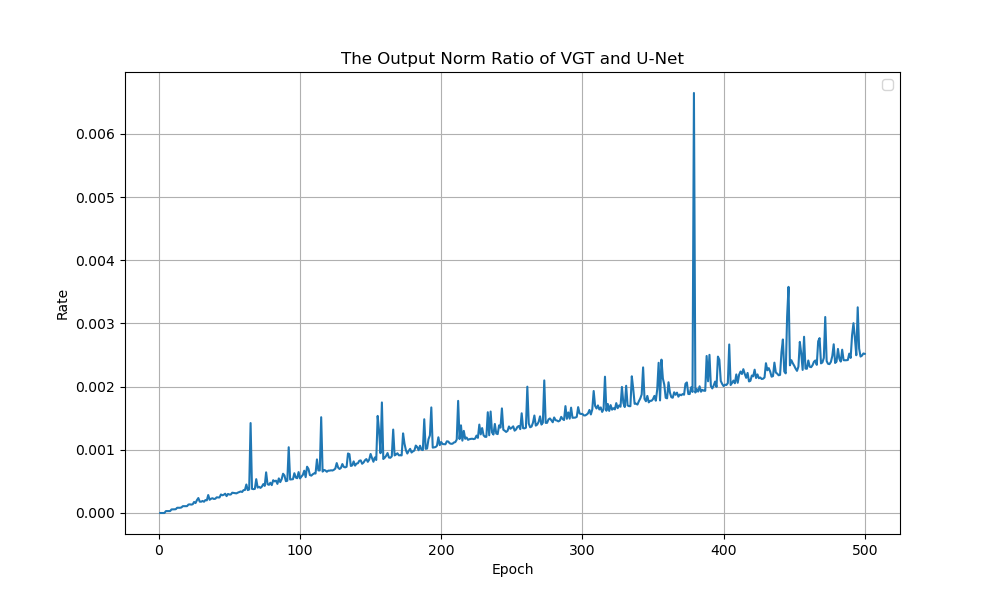} \label{rate}}
  \caption{(a) is the comparison of different versions of VGT. ``EN'' represents the use of a transformer encoder instead of a decoder, which means the mask operation was not included. As the picture shows, VGT-Hyper performs the best while the encoder version of VGT-Hyper performs the worst. For VGT-Pure, the encoder version performs similarly to the decoder version, while the performance of the two versions is between VGT-Hyper and VGT-Hyper-EN. (b) shows the ratio of VGT output and U-Net in the denoising step. The result shows that the norm of VGT output is very tiny compared with the U-Net output, which shows that the output of VGT did not change the original output much while improving the consistency of different frames laterally.}
\end{figure*}

To ensure content consistency with the introduced module, we propose that the module's output should manifest as subtle perturbations, significantly smaller in magnitude than the output of the pre-trained model (referred to as U-Net in this paper). The evolving ratio of perturbation to U-Net output over epochs can be visually observed in Fig.~\ref{rate}. Let the input be denoted as $x\in\mathbb{R}^{B\times H\times W\times F\times C}$, where $B$, $H$, $W$, $F$, and $C$ represent batch, height, width, frame, and channel dimensions respectively. We define the encoder as $\mathcal{E}$, decoder as $\mathcal{D}$, and $\tilde{x}=\mathcal{D}(z)=\mathcal{D}(\mathcal{E}(x))$ signifies the intended outcome. Furthermore, considering the process in the latent space, let $Z$ denote the latent space with $z\in Z$, where $z\in\mathbb{R}^{B\times h\times w\times F\times c}$, where $h,w,c$ are the dimension in the latent space respectively. The diffusion process and U-Net are represented by $\phi$ and $\pi$ respectively. The term $z_{T}$ in the denoising stage signifies the latent variable's evolution over $T$ steps with added noise. A pivotal addition is our module, termed Video Generation Transformer (VGT). Denoted as $\varphi$, it encapsulates an abstract function. The series of denoising autoencoders is represented as $\epsilon_{\theta}(x_{t}, t)$, with $t=1, \ldots, T$, where $t$ corresponds to the specific step in the sequence. Then, we have
\begin{equation}
\label{diff}
    z_{t}=\phi(z_{t-1}, t-1),
\end{equation}
\begin{equation}
    \hat{z}_{T-1}=\pi(z_{T}, T).
\end{equation}
\begin{equation}
    \hat{z}_{t-1}=\pi(\hat{z}_{t}, t),
\end{equation}
where $\hat{z}_{t}$ represents the predicted output of denoising U-Net at the $t$-th step. In fact, we add the perturbation to the U-Net, which aims to capture intrinsic information:
\begin{equation}\label{add}
    \hat{z}_{t-1}^{*}=\pi(\hat{z_{t}}, t)+\varphi(z_{t}, t),
\end{equation}
thus the original object function:
\begin{equation}
    \mathcal{L}_{MSE}:=\mathbb{E}_{\mathcal{E}(x), \epsilon \sim \mathcal{N}(0,1), t}\left[\left\|\epsilon-\epsilon_{\theta}\left(\hat{z}_{t}, t\right)\right\|_{2}^{2}\right],
\end{equation}
can be rewritten as:
\begin{equation}
    \mathcal{L}_{MSE}:=\mathbb{E}_{\mathcal{E}(x), \epsilon \sim \mathcal{N}(0,1), t}\left[\left\|\epsilon-\epsilon_{\theta}\left(\hat{z}_{t}^{*}, t\right)\right\|_{2}^{2}\right].
\end{equation}

For clarity, the intuitive pseudocode is illustrated in the Algrithm~\ref{APLA}. With the adversarial training that enhances the robustness and quality of the generator output, a discriminator is set to receive the predicted noise and noise residuals in the corresponding step. More elaborate discussions are presented in the following sections.
\subsection{Video Generation Transformer (VGT)}
We introduce the proposed VGT, designed as a decoder-only architecture. With self-attention mechanisms, this Transformer efficiently focuses on input features. Compared to pure encoder-based structures like BERT \cite{devlin2019bert}, the pure Transformer decoder architecture \cite{liu2018generating} accommodates more tokens, boosting its processing capacity. The design of the decoder within the Transformer stands out as a unique case in the realm of autoregressive models \cite{dai2015semisupervised}, showcasing potential in unsupervised time series prediction \cite{brown2020language}. Besides, this design empowers the Transformer decoder to extract temporal information from input data. For tasks involving time sequences, the self-attention mechanisms of the Transformer decoder enhance output distribution coherence by extracting contextual insights. Notably, even with incomplete sequence inputs, Transformer decoders promote model diversity \cite{esser2021taming}, thereby bolstering generalization capabilities.

Nonetheless, in previous video generation models employing Transformer architectures, like Video Transformer (VIVIT) \cite{arnab2021vivit}, solely the Transformer encoder is utilized to derive a latent variable. This variable is subsequently fed into another network, often a classifier, to fulfill the downstream task. Consequently, VIVIT lacks the direct capability to generate videos. It is precisely due to this limitation that we introduce the Video Generation Transformer (VGT). Our framework aims to reconstruct or generate videos, leading to the proposal of two distinct VGT variants. The first is a pure Transformer decoder approach, referred to as VGT-pure, while the second combines self-attention with 3D convolution, termed VGT-Hyper.

\begin{table}
\centering
\begin{tabular}{lcc}
\hline
\multicolumn{3}{c}{\textbf{VGT version}}\\ \hline
\multicolumn{1}{l|}{\textbf{Version}} & \multicolumn{1}{c|}{\textbf{PSNR}} &\textbf{Trainable Parameters} \\ \hline
\multicolumn{1}{l|}{VGT-Pure} & \multicolumn{1}{c|}{52.746} & \textbf{60.362M} \\
\multicolumn{1}{l|}{VGT-Hyper} & \multicolumn{1}{c|}{\textbf{58.552}} & 97.136M \\
\multicolumn{1}{l|}{VGT-Pure-EN} & \multicolumn{1}{c|}{54.236} & \textbf{60.362M} \\
\multicolumn{1}{l|}{{VGT-Hyper-EN}} & \multicolumn{1}{c|}{42.736} & 97.136M \\ \hline
\end{tabular}
  \caption{The different version of VGT, while ``EN'' represents the transformer encoder to be used instead of the encoder, which means there is no mask operation. We compare the PSNR of the reconstruction quality, as the input is the single video generated randomly. Meanwhile, we compare the trainable parameters of different VGT, while the VGT-Pure and VGT-Pure-EN own the lowest trainable parameters, for the mask operation did not change the quantities of trainable parameters.}
  \label{vgt-com}
\end{table}

\subsection{VGT-Pure}
The initial model variant is a pure Transformer decoder. We denote the input sequence in the $\ell_{th}$ layer as $\mathbf{z}_{s}^{\ell}$ and $\mathbf{z}_{t}^{\ell}$, where $s$ and $t$ stand for spatial and temporal aspects respectively. Furthermore, we define a token as $\mathbf{z}_{cls, s}^{k, \ell}$, with ``cl'' signifying the class token, and $k$ representing the $k$-th token in the sequence, excluding the class token. In this context, we replace Multi-Headed Self Attention (MAS) with MASKed Multi-Headed Self Attention (MMAS). Likewise, each transformer block encompasses layer normalization (LN). The spatial decoder block can be succinctly represented as follows:
\begin{equation}
\mathbf{z}_{s}^{\ell+1}=\texttt{MMSA}(\texttt{LN}(\mathbf{z}_{s}^{\ell}))+\mathbf{z}_{s}^{\ell},
\end{equation} 
and similarly the temporal decoder block as:
\begin{equation}
\mathbf{z}_{t}^{\ell+1}=\texttt{MMSA}(\texttt{LN}(\mathbf{z}_{t}^{\ell}))+\mathbf{z}_{t}^{\ell}.
\end{equation} 
\begin{table*}[!t]
\centering
\begin{tabular}{lcccc}
\hline
\multicolumn{5}{c}{\textbf{Frame Consistency}}\\ \hline
\multicolumn{1}{l|}{\textbf{Method}} & \multicolumn{1}{c|}{\textbf{CLIP Score}} &  \multicolumn{1}{c|}{\textbf{FVD}}&\multicolumn{1}{c|}{\textbf{IS}} & \textbf{FCI} \\ \hline
\multicolumn{1}{l|}{CogVideo \cite{hong2022cogvideo}} & \multicolumn{1}{c|}{90.64}& \multicolumn{1}{c|}{626} & \multicolumn{1}{c|}{50.46} & \multicolumn{1}{c}{0.2942} \\
\multicolumn{1}{l|}{Plug-and-Play \cite{tumanyan2022plugandplay}} & \multicolumn{1}{c|}{88.89} & \multicolumn{1}{c|}{-} & \multicolumn{1}{c|}{-}& 0.3048 \\
\multicolumn{1}{l|}{VideoGPT \cite{yan2021videogpt}} & \multicolumn{1}{c|}{47.12} & \multicolumn{1}{c|}{-} & \multicolumn{1}{c|}{24.69}& 0.4567 \\
\multicolumn{1}{l|}{DVD-GAN \cite{clark2019adversarial}} & \multicolumn{1}{c|}{48.36} & \multicolumn{1}{c|}{-} & \multicolumn{1}{c|}{27.38}& 0.4012 \\
\multicolumn{1}{l|}{TGANv2 \cite{Saito_2020}} & \multicolumn{1}{c|}{51.26} & \multicolumn{1}{c|}{1209} & \multicolumn{1}{c|}{28.87}& 0.3538 \\
\multicolumn{1}{l|}{MoCoGAN-HD \cite{tian2021good}} & \multicolumn{1}{c|}{77.64} & \multicolumn{1}{c|}{838} & \multicolumn{1}{c|}{32.36}& 0.3156 \\
\multicolumn{1}{l|}{DIGAN \cite{yu2022generating}} & \multicolumn{1}{c|}{81.26} & \multicolumn{1}{c|}{655} & \multicolumn{1}{c|}{29.71}& 0.3012 \\
\multicolumn{1}{l|}{TATS-base \cite{ge2022long}} & \multicolumn{1}{c|}{90.65} & \multicolumn{1}{c|}{\textbf{332}} & \multicolumn{1}{c|}{\textbf{79.28}}& 0.2647 \\
\multicolumn{1}{l|}{Tune-A-Video \cite{wu2022tune}} & \multicolumn{1}{c|}{92.40} & \multicolumn{1}{c|}{864}& \multicolumn{1}{c|}{64.12}& 0.2716 \\
\multicolumn{1}{l|}{SAVE \cite{karim2023save}} & \multicolumn{1}{c|}{94.81} & \multicolumn{1}{c|}{-}& \multicolumn{1}{c|}{-}& 0.2628 \\
\multicolumn{1}{l|}{\textbf{APLA (Ours)}} & \multicolumn{1}{c|}{\textbf{96.21}} & \multicolumn{1}{c|}{512}& \multicolumn{1}{c|}{71.26}& \textbf{0.2576} \\ \hline
\end{tabular}
  \caption{Quantitative comparison with evaluated baselines by measuring content consistency with text prompt using CLIP Score ($\uparrow$), FVD ($\downarrow$), IS ($\uparrow$) and FCI ($\downarrow$), which measures how many users prefer each model. FCI ($\downarrow$) is the metric that computes the optical flow difference of the adjacent frame. The lower, the better.}
  \label{tab:compare}
\end{table*}

We consider $L$ layers of spatial decoder at all, and the $\mathbf{z}_{t}^{L}$ can be represented as:
\begin{equation}\label{...}
\mathbf{z}_{s}^{L}=\left[\mathbf{z}_{cls, s}^{L}, \mathbf{z}_{s}^{1,L}, \mathbf{z}_{s}^{2,L}, \ldots, \mathbf{z}_{s}^{F,L}\right]+\mathbf{p},
\end{equation}
where $\mathbf{p}$ denotes the positional embedding, and $\mathbf{z}$ is divided into tokens. Among these tokens, the first one is $\mathbf{z}_{cls,s} \in \mathbb{R}^{F\times (H_{patch}\times W_{patch}) \times L}$, serving as a compact feature commonly used for categorical embedding. Here, $H_{patch}$ represents the patch's height, and $W_{patch}$ is the width of a patch. When we split $\mathbf{z}_{cls,s}$ along the temporal frame dimension, we obtain individual tokens $\mathbf{z}_{cls,s}^{1,L}, \mathbf{z}_{cls,s}^{2,L}, \ldots, \mathbf{z}_{cls,s}^{F,L}$, where $\mathbf{z}_{cls,s}^{i,L}\in\mathbb{R}^{1\times (H_{patch}\times W_{patch}) \times L}$ for $i=1, 2, \ldots, F$. Simultaneously, we maintain the collection $\left[\mathbf{z}_{s}^{1,L}, \mathbf{z}_{s}^{2,L}, \ldots, \mathbf{z}_{s}^{F,L}\right]$ as $\mathbf{z}_{s}^{C,L}$, facilitating skip connections. This arrangement leads us to use $\mathbf{z}_{cls,s}^{L}$ as the input for the temporal decoder, setting it apart from the rest. This ultimately yields the expression for $\mathbf{z}_{t}^{1}$:
\begin{equation}\label{...}
\mathbf{z}_{t}^{1}=\left[\mathbf{z}_{cls,t}^{1}, \mathbf{D} \mathbf{z}_{cls,s}^{1,L}, \mathbf{D} \mathbf{z}_{cls,s}^{2,L}, \ldots, \mathbf{D} \mathbf{z}_{cls,s}^{F,L}\right]+\mathbf{p},
\end{equation}
where $\mathbf{D}$ is the decoder block. Suppose we have $M$ layers temporal decoder in total. Similarly, the output of $M_{th}$ can be written as:
\begin{equation}\label{...}
\mathbf{z}_{t}^{M}=\left[\mathbf{z}_{cls, t}^{M}, \mathbf{z}_{cls,t}^{1,M}, \mathbf{z}_{cls,t}^{2,M}, \ldots, \mathbf{z}_{cls,t}^{F,M}\right]+\mathbf{p},
\end{equation}
Similarly, we denote $\left[\mathbf{z}_{cls,t}^{1,M}, \mathbf{z}_{cls,t}^{2,M}, \ldots, \mathbf{z}_{cls,t}^{F,M}\right]$ as $\mathbf{z}_{t}^{C,M}$, then the output of VGT-Pure can be written as:
\begin{equation}\label{equ11}
\mathbf{y} = \mathbf{z}_{s}^{C,L} \odot \mathbf{z}_{t}^{C,M},
\end{equation}
\begin{equation}\label{...}
\hat{\mathbf{z}_{t}} = \texttt{MLP}(\mathbf{y}),
\end{equation}
where $\odot$ represents the Hadamard product, MLP represents multilayer perceptron and $t$ is $t$-th step.
we get $\hat{\mathbf{z}_{t}}\in \mathbb{R}^{B\times F\times (H_{patch}\times W_{patch}) \times (C \times P \times P)}$, then we rearrange the $\hat{\mathbf{z}_{t}}$ into $\varphi(z_{t})$, where $\varphi(z_{t})\in R^{B\times F \times C\times (H_{patch}\times P)\times (W_{patch}\times P)}$ and $\varphi(\cdot)$ is the function representation of VGT-Pure while $z_{t}$ is the input of VGT-Pure, and $(H_{patch}\times P) = H$ and $(W_{patch}\times P) = H$.

\subsection{VGT-Hyper}
In this section, we introduce the second variant of VGT, named VGT-Hyper, which leverages 3D convolution\cite{tran2015learning}. Particularly, in Eq.~\ref{equ11}, rather than employing element-wise multiplication, we opt for 3D convolution. We represent the convolution block with the matrix $\mathbf{M}$, leading to the following expression:
\begin{equation}\label{...}
\mathbf{y^{*}} =  \mathbf{M}\mathbf{z}_{t}^{M},
\end{equation}
\begin{equation}\label{...}
\hat{\mathbf{z}_{t}} = \texttt{MLP}(\mathbf{y^{*}}).
\end{equation}

Unlike VGT-pure, VGT-Hyper demonstrates superior performance in the reconstruction task, as indicated in Tab.~\ref{vgt-com}, all the while maintaining a higher number of trainable parameters. VGT-Hyper capitalizes on the benefits inherent in a transformer decoder, underscoring the efficacy of the mask operation for time series tasks, depicted in Fig.~\ref{VGT_COM}.

\subsection{Hyper-Loss for Latent Noise Fitting}
Recognizing the limitations of Mean Squared Error (MSE) for certain generative tasks \cite{zhang2018unreasonable}, we introduce a novel loss function tailored for video generation. We adopt a perceptual loss approach within the diffusion model, akin to prior studies \cite{Lugmayr_2022_CVPR}. In detail, we formalize $\ell_1$ loss and perceptual loss separately as follows:
\begin{equation}
    \mathcal{L}_{L1}:=\mathbb{E}_{\mathcal{E}(x), \epsilon \sim \mathcal{N}(0,1), t}\left[\left\|\epsilon-\epsilon_{\theta}\left(z_{t}^{*}, t\right)\right\|_{1}\right],
\end{equation}
and
\begin{equation}
    \mathcal{L}_{per}:=\mathbb{E}_{\mathcal{E}(x), \epsilon \sim \mathcal{N}(0,1), t}\left[dist_{per}(\epsilon, \epsilon_{\theta}\left(z_{t}^{*}, t\right))\right].
\end{equation}

Expanding upon this, we incorporate a hyper-loss that encompasses the weighted combination of Mean Squared Error (MSE), $\ell_1$ loss, and perceptual loss. The $\ell_1$ loss serves as a regularization term to promote sparsity in the solution, while the perceptual loss encourages the model to generate more photorealistic images. This concept is illustrated as follows:
\begin{equation}\label{...}
\mathcal{L}_{hyper}:=\alpha*\mathcal{L}_{MSE}+\beta*\mathcal{L}_{L1}+\gamma*\mathcal{L}_{per},
\end{equation}
where $\alpha$, $\beta$ and $\gamma$ represent the weights.

\begin{table*}[!t]
    
  \centering
     \begin{tabular}{lcccc}
    \hline
    \multicolumn{5}{c}{\textbf{Frame Consistency} } \\ \hline
    \multicolumn{1}{l|}{\textbf{Method}} & \multicolumn{1}{c|}{\textbf{CLIP Score}}& \multicolumn{1}{c|}{\textbf{FCI}} & \multicolumn{1}{c|}{\textbf{CLIP Score (1500 epochs)}}& \textbf{FCI (1500 epochs)} \\ \hline
    \multicolumn{1}{l|}{Full Model (ours)} & \multicolumn{1}{c|}{\textbf{96.21}}& \multicolumn{1}{c|}{0.2764} & \multicolumn{1}{c|}{\textbf{96.76}} & 0.2470\\
    \multicolumn{1}{l|}{w\textbackslash o Discriminator} & \multicolumn{1}{c|}{94.42}& \multicolumn{1}{c|}{0.2714} & \multicolumn{1}{c|}{93.70} & 0.2178\\
    \multicolumn{1}{l|}{w\textbackslash o VGT\&Discriminator} & \multicolumn{1}{c|}{91.44} & \multicolumn{1}{c|}{\textbf{0.1918}} & \multicolumn{1}{c|}{96.13} & 0.2655\\
    \multicolumn{1}{l|}{w\textbackslash o Hyper-Loss\&Discriminator} & \multicolumn{1}{c|}{93.97}& \multicolumn{1}{c|}{0.2476} & \multicolumn{1}{c|}{93.06} & 0.2588\\
    \multicolumn{1}{l|}{w\textbackslash o VGT\&Hyper-Loss} & \multicolumn{1}{c|}{94.83}& \multicolumn{1}{c|}{0.2534} & \multicolumn{1}{c|}{96.38} & \textbf{0.2172}\\ \hline
    \end{tabular}
\caption{Ablation studies on APLA's different components. Compared via CLIP score and FCI on each model variation trained with 750 epochs(default) and 1500 epochs respectively.}
  \label{aba-study}
\end{table*}
\subsection{Adversarial Training with 1$\times$1 Convolution}
\label{adtrain}
In our approach, we view adversarial training as a valuable form of regularization. For video generation, the distinction between integrating Generative Adversarial Networks (GANs) and employing perceptual loss lies in their treatment of temporal information. Perceptual loss is primarily concerned with the structural attributes of individual frames, whereas reconstruction loss focuses on pixel-level closeness. In contrast, GAN loss centers around maintaining consistency across frames, promoting temporal coherence. Besides, it is important to recognize that GANs have a propensity for capturing global information by treating all frames holistically. This approach leads to an enhancement in video quality through adversarial training. The discriminator, in the APLA, receives the output of the generator, namely the predicted noise to compare with the noise residual, which is obtained in the diffusion process. More concretely, the diffusion process $T$ steps, while the denoising process aims to inverse this process, which predicts the noise residual, the difference of the $t$-th step and $t$-$1$-th step, namely the adding noise. In the denoising process, for instance, the generator predicts the $t$-th step noise residual, and then the discriminator receives the corresponding $t$-th step noise in the diffusion process, aiming to decline the distance of two noise distributions (noise residual and predicted noise residual).
 
The proposed discriminator structure is streamlined, comprising only a 1$\times$1 convolutional layer. This kernel comprehensively considers frame positional data, aiding temporal similarity extraction. Denoting the discriminator as $D(\cdot)$ and with $x \sim p(x)$, we arrive at a min-max problem:
\begin{equation}\label{...}
\mathop{\max}_{D}\mathop{\min}_{G} 
\mathbb{E}_{{x} \sim p(x)}
\left[\log D_{G}^{*}({x})\right]
+\mathbb{E}_{{x} \sim p_{g}}
\left[\log \left(1-D_{G}^{*}
({x})\right)\right],
\end{equation}
where $G$ represents the generator, which is the united block of U-Net and VGT in this paper, while $D$ represents the discriminator. Let the generation loss of the generator, in this paper, which is the fusion of U-Net and VGT, as $L_{g}$, where $L_{g} = \mathbb{E}_{{x} \sim p_(x)}\left[\log D_{G}^{*}({x})\right]+\mathbb{E}_{{x} \sim p_(x)}\left[\log \left(1-D_{G}^{*}({x})\right)\right]$. Hence, the final optimization objective is
\begin{equation}\label{final}
\mathop{\min}_{\theta}\mathop{\max}_{G}\mathcal{L}_{hyper}+\lambda L_{g},
\end{equation}
where $\lambda$ is a coefficient to adjust the performance and $\theta$ is the parameters of the network.


\section{Experiments}
\label{others}

\subsection{Implementation Details}
We build the model based on Tune-A-Video \cite{wu2022tune, elmquist2022art} and utilize the pre-trained weights of the Stable Diffusion \cite{rombach2022high}. We uniformly sample 24 frames at a resolution of 512$\times$512 from the input video and train the models for 750 steps and 1500 steps with a learning rate of 3e-5 and a batch size of 1. For the hyperparameter induced by loss, we set $\alpha$ of 0.5, $\beta$ of 0.2, $\gamma$ of 0.1, and $\lambda$ of 0.5. Due to the limited CUDA memory, we chose the VGT-pure only for the following experiment. During inference, we employ the DDIM sampler \cite{song2020denoising} with classifier-free guidance \cite{ho2022classifier} in our experiments. For a single video, it takes about 90 minutes for 1500 steps and approximately 1 minute (while Tune-A-Video takes 60 minutes) for sampling on an NVIDIA 3090 GPU.

{\bf Dataset:} To evaluate our approach, we use representative videos taken from the DAVIS dataset \cite{pont20172017}. During fine-tuning, we just train the model on a single video. The video descriptions and captions automatically used an off-the-shelf captioning model \cite{li2023blip}, which is regarded as the default prompt of our video. 

{\bf Qualitative Results:} A visual comparison is presented between our approach and the baseline in Fig.~\ref{reco}, focusing on reconstructing tasks using the same prompt. We observe that Tune-A-Video does not reconstruct the details in each frame well. Specifically, the sled visibly splits into two pieces in some frames. Our model provides more stable and smooth results compared with the Tune-A-Video. Also, readers can find more results obtained by our method in Fig. \ref{result}. 

{\bf Quantitative Results:} Quantitative assessment is conducted, as depicted in Tab.~\ref{tab:compare}. Regarding content consistency, we evaluate the CLIP score \cite{radford2021learning} across all generated frames by employing average cosine measurements. This metric serves as an indicator of the semantic coherence within the generated videos. In terms of frame consistency, we employ the flow consistency index (FCI) for comparison. Unlike\cite{9150870}, we directly compute the flow field between consecutive frames, independent of the input video. Specifically, we determine the optical flow field between two adjacent frames, assess alterations in the local optical flow field concerning each pixel's value and its domain, and subsequently average all the computed changes. The results underscore that, in comparison to our baseline model, enhancements are observed in both content consistency and frame consistency.

\subsection{Ablation Studies} We conduct ablation studies to assess the importance of different components of ALVA, as shown in Tab.~\ref{aba-study}. The proposed full APLA model performs the best considering content consistency and frame consistency together. Without some components, APLA's performance degrades but still performs better than Tune-A-Video. We also discuss the influence of the number of epochs. From observing Tab.~\ref{aba-study}, we see that with the epoch increasing, the quantitative score is increasing. However, from the visual result, we see that too many epochs can cause overfitting, which destroys the result influenced by the prompt. Without the Discriminator, as the number of epochs increases, it is easy to fall into local minima which decreases the CLIP score and FCI. For w\textbackslash o VGT\&Discriminator, although the final FCI is decent, it cannot retain the semantic consistency and needs too many epochs to reach a good result. For w\textbackslash o Hyper-Loss\&Discriminator, the single VGT can just reach a normal level and it is hard for it to approach a better score because of the limitation of convergence. As w\textbackslash o VGT\&Hyper-Loss, the model performance is close to w\textbackslash o Discriminator and even better. However, it still needs too many epochs to reach such a good result.


\section{Conclusion} In this study, we introduce APLA, which includes a compact module for capturing intrinsic or temporal information, and the novel VGT architecture, a pure transformer decoder similar to GPT. To fortify the robustness and quality of our APLA model, we employ adversarial training during its training process. Through experiments, our model achieves state-of-the-art performance in video reconstruction and videos from textual prompts (T2V).


\medskip

\bibliography{ref}

\end{document}